\documentclass[11pt,a4paper]{article}
\usepackage[hyperref]{naaclhlt2018}
\usepackage{times}
\usepackage{latexsym}

\usepackage{url}

\usepackage{graphicx}
\usepackage{multirow}

\aclfinalcopy % Uncomment this line for all SemEval submissions

\title{NELEC at SemEval-2019 Task 3: Think Twice Before Going Deep}

\author{Parag Agrawal\thanks{\ \ Equal contribution, order determined by coin toss} \\
  Microsoft \\
  {\tt paragag@microsoft.com} \\\And
  Anshuman Suri\footnotemark[1]  \\
  Microsoft \\
  {\tt ansuri@microsoft.com} \\}

\date{}

\begin{document}
\maketitle
\begin{abstract}
  Existing Machine Learning techniques yield close to human performance on text-based classification tasks. However, the presence of multi-modal noise in chat data such as emoticons, slang, spelling mistakes, code-mixed data, etc. makes existing deep-learning solutions perform poorly. The inability of deep-learning systems to robustly capture these covariates puts a cap on their performance. We propose \textbf{NELEC} : \textbf{Ne}ural and \textbf{Le}xical \textbf{C}ombiner, a system which elegantly combines textual and deep-learning based methods for sentiment classification. We evaluate our system as part of the third task of 'Contextual Emotion Detection in Text' as part of SemEval-2019~\cite{SemEval2019Task3}. Our system performs significantly better than the baseline, as well as our deep-learning model benchmarks. It achieved a micro-averaged $F_{1}$ score of 0.7765, ranking $3^{rd}$ on the test-set leader-board. Our code is available at ~\url{https://github.com/iamgroot42/nelec}
\end{abstract}

\section{Introduction}
Sentiment analysis of textual data: Twitter data~\cite{kouloumpis2011twitter, pak2010twitter}, movie reviews~\cite{thet2010aspect}, and product reviews~\cite{pang2008opinion}, is perhaps the most extensively explored problem, with a plethora of research to tackle it. Novel systems utilise deep learning architectures to achieve near-human performance on clean, well-formatted data. However, sentiment classification of chat data is significantly challenging. The presence of spelling errors, slang, emoticons, code-mixing, style of writing and abbreviations makes it significantly harder for existing deep-learning models to work on such data. \par
\begin{figure*}[t]
\begin{center}
    \includegraphics[width=1.0\linewidth]{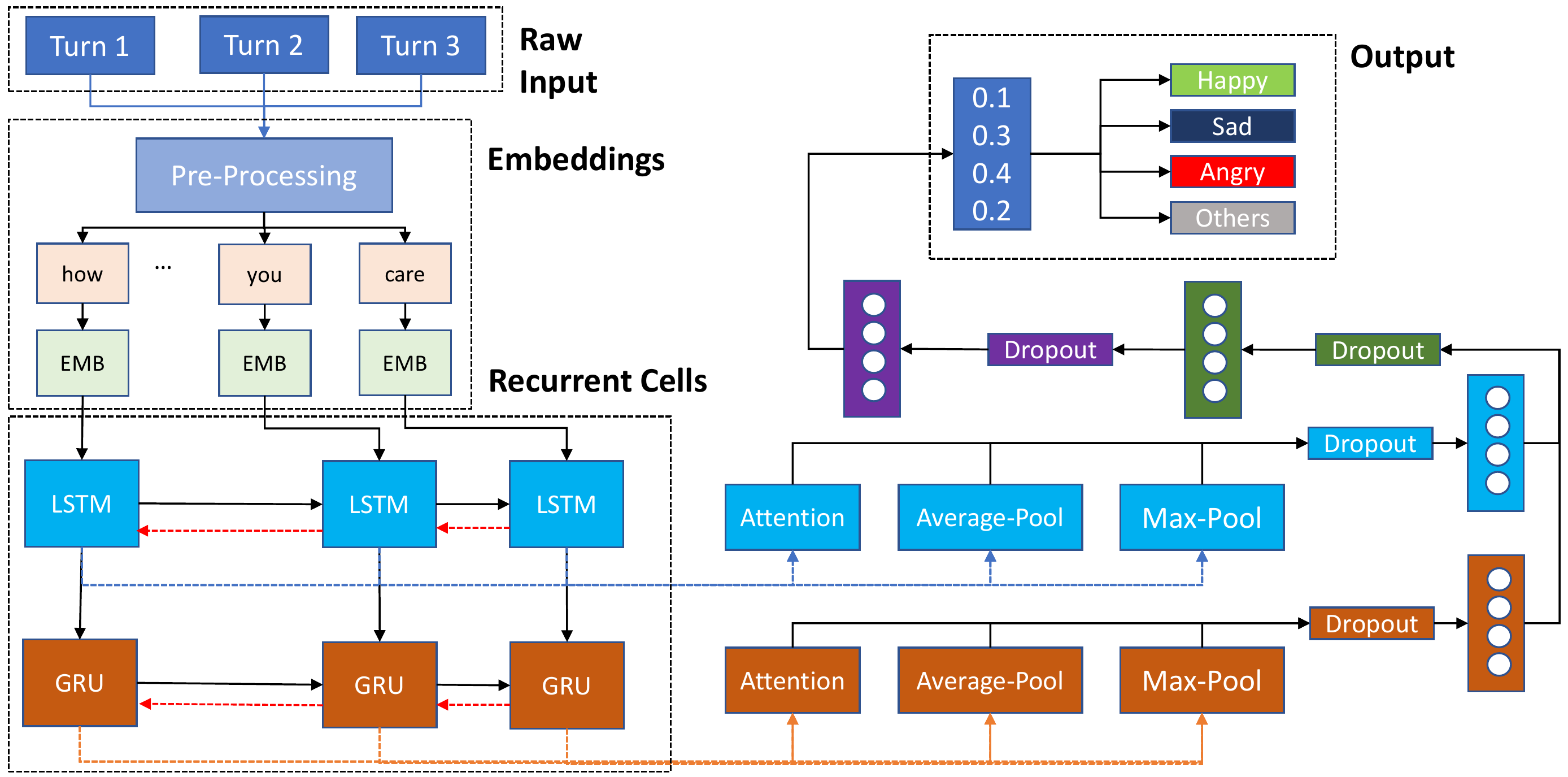}
\end{center}
  \caption{System diagram of the Deep-Learning model described in Section~\ref{deep_description}.}
\label{fig:deep_architecture}
\end{figure*}
Literature dealing with this problem comprises a wide range of approaches: from hand-crafted features to end-to-end deep-learning methods. Some rule-learning based methods use keyword-based analysis~\cite{ko2000automatic} and part-of-speech tagging~\cite{agarwal2011sentiment}. These procedures require extensive human-involvement for identifying keywords and designing rules and are thus not scalable. \par
Non-neural machine-learning methods utilize feature extraction algorithms like \textit{n}-grams and Tf-Idf vectors, coupled with classification algorithms like Naive Bayes~\cite{pang2002thumbs}, Decision Trees~\cite{bilal2016sentiment}, SVM~\cite{moraes2013document}. These approaches perform significantly better than rule-based approaches but fail to capture context well, since they ignore the order of words in text sequences. \par
\begin{table}[h!]
\centering
 \begin{tabular}{||c c c c||} 
 \hline
 \textbf{Statistic} & \textbf{Train} & \textbf{Dev} & \textbf{Test} \\
 \hline\hline
 Emojis (\%) & 17.6 & 11.1 & 12.5 \\ 
 \hline
 OOV (\%) & 3.7 & 4.9 & 4.9 \\
 OOV(processed) (\%) & 2.1 & 1.5 & 1.8 \\
 \hline
 Avg.Length & 13.6 & 12.7 & 12.7 \\
 Avg.Length(processed) & 15.7 & 15.3 & 15.2 \\
 \hline
 Happy emotion (\%) & 14.1 & 5.2 & 5.2 \\
 Sad emotion (\%) & 18.1 & 4.5 & 4.5 \\ 
 Angry emotion (\%) & 18.3 & 5.4 & 5.4 \\ 
 \hline
 \end{tabular}
 \caption{Some statistics for the given training, development and test sets.}
 \label{table:data_stats}
\end{table}
Neural, deep-learning based approaches use architectures such as variations of recurrent models: GRU~\cite{chung2014empirical}, LSTM~\cite{hochreiter1997long}, BiLSTM~\cite{schuster1997bidirectional} and Convolutional models~\cite{mundra2017fine}, performing significantly better than other machine-learning techniques. Their ability to generalise and capture context over long sequences makes them a popular choice for text classification tasks. \par
We propose \textbf{NELEC}, a novel system specifically designed for sentiment classification. We combine lexical and neural features for sentiment classification, followed by class-specific thresholds for better labelling. Our system yields an $F_{1}$ score of 0.7765 on the test-set of Task 3 of Sem-Eval 2019.

\section{System Description}

\subsection{Deep Learning Model} \label{deep_description}
We experiment with a two-layer, recurrent, deep-learning model with skip connections, bidirectional cells and attention (Figure~\ref{fig:deep_architecture}). We trained our model for 100 epochs with Cyclic Learning Rate~\cite{smith2017cyclical} scheduling. This model outperforms the baseline by a significant margin. An in-depth analysis of the cases where it fails reveals its shortcomings (along with that of a deep-learning model in general): it is not robust to misspellings and cannot capture the meaning of out-of-vocabulary words robustly. Even though pre-trained embeddings are available for most words, the context with which they are used in chat may vary from the corpora they were trained on, thus lowering their usability.
\subsection{NELEC : Neural and Lexical Combiner} \label{system_description}
Since neural features have a lot of shortcomings, we shift our focus to lexical features. Using a combination of both lexical (\textit{n}-gram features, etc.) and neural features (scores from neural classifiers), we trained a standard Light-GBM ~\cite{ke2017lightgbm} Model for 100 iterations, with feature sub-sampling of 0.7 and data sub-sampling of 0.7 using bagging with a frequency of 1.0. We use $10^{-2} * \|weights\|_{2}$ as regularization. We also experimented with a logistic regression model, but it had a significant drop in performance for the 'happy' and 'angry' classes (Table~\ref{table:f1_results}). The total number of features used is 9270, out of which 9189 are sparse. The features we use in our model are described in the sections below:

\subsubsection{Turn Wise Word \textit{n}-Grams} \label{turnwisewordgram}
Word level bi-grams and tri-grams (skip 1). These help capture patterns like ``am happy" and automatically handles unseen data such as ''am very happy" or "am so happy" because of the skip word. We take the term frequencies of these \textit{n}-Grams as features. Word Grams not$\vert$good, hate, no$\vert$one had the highest feature gains.

\subsubsection{Turn Wise Char \textit{n}-Grams} \label{turnwisechargram}
Character level bi-grams and tri-grams. This feature helps capture character-level trends such as ``haha" (and its variants), as well as emoticons. It helps with misspellings and makes the system robust to variants of several words like ``haha". h$\vert$a$\vert$h, w$\vert$o$\vert$w had one of the highest feature gains.

\subsubsection{Valence Arousal Dominance} \label{valencearousal}
We used Valence-Arousal-Dominance data~\cite{mohammad2018obtaining} in the following manner:
\begin{enumerate}
    \item Mean of Valence and Arousal values, along with turn-wise Maximum Dominance value for all words. Turn 3 Arousal for maximum dominant word had the highest feature gain.
    \item Turn-wise mean of Valence, Arousal and Dominance values.
\end{enumerate}

\subsubsection{Emotion Intensity}\label{emotionintensity}
We use EmoLex~\cite{Mohammad2010EmotionsEB}, which associates words to eight emotions and two sentiments. For each turn, we obtain the number of words having specific emotions and sentiment and use it as a feature.
% Turn 3 Sad word count yielded one of the highest feature gains.

\begin{table}[h]
\centering
 \begin{tabular}{||c |c |c |c |c||} 
 \hline
 \multirow{2}{*}{Model} & \multicolumn{4}{|c||}{$F_{1}$} \\
 \cline{2-5}
  & happy & sad & angry & $\mu_{avg}$\\
 \hline  
 \hline\hline
 \multicolumn{5}{|c||}{Without Data Pre-Processing} \\
 \hline
 Deep & .5863 & .5977 & .6485 & \underline{.6123} \\ 
 \hline
 NELEC & .7382 & .8047 & .7873 & \textbf{.7765}\\
 Logistic & .6712 & .7642 & .7151 & .7154 \\
 \hline
 Baseline & .5461 & .6149 & .5945 & .5861\\
 \hline
 \multicolumn{5}{|c||}{With Data Pre-Processing} \\
 \hline
 Deep & .5710 & .6630 & .7350 & \underline{.6651} \\ 
 \hline
 NELEC &  .7324 & .8015  & .7878 & \textbf{.7736}  \\
 Logistic & .6782 & .7680 & .7120 & .7177 \\
 \hline
 Baseline & .5797 & .5973 & .6241 & .6024 \\
 \hline
 \end{tabular}
 \caption{Class-wise and micro-averaged $F_{1}$ scores for NELEC, our deep-learning model and existing baseline.}
 \label{table:f1_results}
\end{table}

\subsubsection{Neural Features} \label{neuralfeatures}
We used scores obtained by utilizing available pre-trained classifiers features:
\begin{enumerate}
    \item Scores obtained by running conversations through a Sentiment Classifier trained on Twitter Data using SSWE embeddings~\cite{tang2014learning}. 
    \item Signals from Adult and Offensive Classifiers~\cite{yenala2017deep}, obtained via the Text Moderation API by Microsoft Cognitive Services. As observed in Table \ref{table:f1_results}, this helps in 'Anger' detection. \footnote{https://docs.microsoft.com/en-in/azure/cognitive-services/content-moderator/text-moderation-api}
\end{enumerate}

\subsection{Lexical Count Features} \label{Lexical Features}
Lastly, we used certain count features such as the number of interrogation marks, exclamation marks, uppercase letters, the total number of words and letters for each turn. These features were observed to be very helpful while detecting anger and happiness.

\section{Data Preparation}

The training, development and test sets consist of 30160, 2755 and 5509 examples respectively. The final model is trained on the combined training and development set. For each instance, one of four class labels: $\{$happy, angry, sad, other$\}$, is provided. Table~\ref{table:data_stats} provides some statistics for the given dataset.
\begin{table*}[t]
\centering
 \begin{tabular}{||c |c |c |c | c| c| c||} 
 \hline
 Feature Dropped & Features (\#) & $F_{1_{\mu_{avg}}}$ & Angry $F_{1}$ & Sad $F_{1}$ & Happy $F_{1}$ & $F_{1_{\mu_{avg}}}$ gain\\
 \hline
 Word \textit{n}-grams & 4565 & .7355 & .7373 & .7723 & .6995 & .0410\\
 Character \textit{n}-grams & 4624 & .6067 & .6271 & .6168 & .5749 & \textbf{.1698}\\
 Valence-Arousal & 15 & .7444 & .7125 & .7426 & .7160 & .0321\\
 Word-emotion Classifier & 30 & .7537 & .7584 & .7739 & .7301 & .0228\\
 Pre-Built Classifier & 9 & .7524 & .7373 & .7756 & .7481 & .0241\\
 Lexical Count Features & 27 & .7654 & .7751 & .8015 & .7217 & .0111\\
 \hline
 Turn 1 (All Features) & 2578 & .7417 & .7173 & .7716 & .7106 & .0348\\
 Turn 2 (All Features) & 3873 & .7642 & .7719 & .8015 & .7217 & .0123\\
 Turn 1 \& 2 (All Features) & 6451 & .7191 & .7304 & .7539 & .6750 & \underline{.0574}\\
 \hline
 \end{tabular}
 \caption{Micro-averaged $F_{1}$ scores when all features apart from these (per row) are dropped. $F_{1}$ gain here refers to the gain when using the feature mentioned, as opposed to dropping it.}
 \label{table:ablation}
\end{table*}

We concatenate all three turns per conversation. For the Deep-Learning approach, a special $\langle eos \rangle$ token is inserted in between these turn-conversations.

\subsection{Pre-processing for NELEC} \label{lexical_preprocess}

\begin{enumerate}
    \item \textbf{Lemmatization}: Contrary to intuition, using lemmatization decreased the final performance of our model. Further analysis suggests that emotion is highly sensitive to exact words: information captured by the word ``hate" and ``hated" are very different, even though a lemmatization system would reduce them to the same word, and similarly for ``happy" versus ``happiest". Using lemmatization drops the system's $F_{1}$ score by 0.0092.
    \item \textbf{WordNet for Synonyms}: We also tried using synonyms for nouns using the Wordnet Graph~\cite{miller1998wordnet}. However, a similar issue plagues this approach. For instance ``dog", ``doggie" and ``puppy" are all synonyms, but they do not express the same kind of emotion: words like ``puppy" convey much more positive emotion. Using Wordnet drops the system's $F_{1}$ score by 0.0023. 
    \item \textbf{Normalization}: We try word tokenization and normalization by removing diacritics, numbers, stop-words, question marks etc.  However, this also drops the $F_{1}$ score by 0.0046.
\end{enumerate}

Character \textit{n}-gram features can handle lemmatization as well as misspellings for most of the cases without discarding any additional information. Finally, we only lower-cased the sentences.

\subsection{Pre-processing for Deep-Learning based Approach} \label{deep_preprocess}
We use pre-trained GloVe~\cite{pennington2014glove} embeddings. Some observations are:
\begin{itemize}
    \item \textbf{Emoticons}: Around 15\% of all conversations includes at least one emoticon. We use embeddings from a pre-trained emoji2vec~\cite{eisner2016emoji2vec} model to handle emoticons.
    \item \textbf{Words with repeated characters}:  This trend is common for chat-data. For example, ``heelloo", ``ookayy". We design specific regular expressions to handle such variations.
    \item \textbf{Abbreviations and slang}: tokens such as ``idk", ``irl" are converted to their full forms.
\end{itemize}

\section{Experiments}

To ascertain the novelty of our system, we report both class-wise and micro-averaged $F_{1}$ scores on the test set. We also compare our performance with the benchmarks provided by the contest organizers ~\cite{chatterjee2019understanding}.

As mentioned in Section~\ref{deep_preprocess}, data pre-processing on deep-learning models leads to significant performance gains, while leading to a drop in performance when using \textbf{NELEC}. \textbf{NELEC} outperforms both the baseline and our deep model by a considerable margin (Table~\ref{table:f1_results}).

\subsection{Ablation Study}

To analyze the usefulness of all features used by \textbf{NELEC}, we perform hold-one-out experiments on its features (Section~\ref{system_description}). Results are reported in Table~\ref{table:ablation}. There is a noticeable gain for most of the features, with character \textit{n}-grams observing the maximum gain among them all.

One of the most intriguing patterns observed is the ease with which they detect sad emotion and an equal difficulty in detecting happiness.
\begin{itemize}
    \item Words like ``haha" and ``okay" have several forms which all convey different magnitudes of emotion. While lemmatising such words, there is a significant loss of information.
    \item Most of the conversations labelled sad have easy-to-recognize signals such as negative emoticons, keywords like ``lonely", which make detection easy. On the other hand, differentiating \textit{happy} and \textit{others} is non-trivial.
    \item Not using the second turn, along with its associated features, leads to a negligible drop in $F_{1}$ performance. This observation highlights the importance of the first user (in data) in analyzing sentiment. Moreover, we can utilize this information to make the feature set even smaller, making the model smaller and faster.
\end{itemize}

\section{Conclusion}
We propose a deep neural architecture to solve the problem of emotion detection in conversations from chat data. Although it outperforms the existing baseline, its performance is not satisfactory. To better capture lexical features and make the model robust to misspellings, abbreviations, emoticons, etc., we propose \textbf{NELEC}, a \textbf{Ne}ural and \textbf{Le}xical \textbf{C}ombiner. Our model utilises lexical features, along with signals from pre-trained neural models for sentiment and adult-offensive classification to boost performance. Our system performs at par with the existing state of the art, yielding a micro-averaged $F_{1}$ score of 0.7765 on the test set, ranking $3^{rd}$ on the test-set leader-board.

\bibliography{semeval2018}
\bibliographystyle{acl_natbib}

\end{document}